\documentclass[letterpaper, 10 pt, conference]{ieeeconf}  

\IEEEoverridecommandlockouts                              

\usepackage{graphicx}%
\usepackage{multirow}%
\usepackage{amsmath,amssymb,amsfonts}%
\usepackage{mathrsfs}%
\usepackage{xcolor}%
\usepackage{textcomp}%
\usepackage{manyfoot}%
\usepackage{booktabs}%
\usepackage{algorithm}%
\usepackage{algorithmicx}%
\usepackage{algpseudocode}%
\usepackage{listings}%

\usepackage{verbatim}
\usepackage{soul}
\usepackage[hidelinks]{hyperref}
\usepackage{xcolor}
\usepackage{cite}

\title{\LARGE \bf
	How Fly Neural Perception Mechanisms Enhance Visuomotor Control of Micro Robots
}

\author{
	Renyuan Liu, Haoting Zhou, Chuankai Fang, and Qinbing Fu$^{*}$ 
	\thanks{
		This research was supported by the National Natural Science Foundation of China under Grant No. 62376063 and 12571558.
	}
	\thanks{All authors are with the Machine Life and Intelligence Research Centre, School of Mathematics and Information Science, Guangzhou University, 510006, China}%
	\thanks{Corresponding author: Qinbing Fu
		{\tt\small qifu@gzhu.edu.cn}}%
}

\begin{document}

\maketitle
\thispagestyle{empty}
\pagestyle{empty}

\begin{abstract}

Anyone who has tried to swat a fly has likely been frustrated by its remarkable agility. 
This ability stems from its visual neural perception system, particularly the collision-selective neurons within its small brain. 
For autonomous robots operating in complex and unfamiliar environments, achieving similar agility is highly desirable but often constrained by the trade-off between computational cost and performance. 
In this context, insect-inspired intelligence offers a parsimonious route to low-power, computationally efficient frameworks. 
In this paper, we propose an attention-driven visuomotor control strategy inspired by a specific class of fly visual projection neurons—the lobula plate/lobula column type-2 (LPLC2)—and their associated escape behaviors. 
To our knowledge, this represents the first embodiment of an LPLC2 neural model in the embedded vision of a physical mobile robot, enabling collision perception and reactive evasion. 
The model was simplified and optimized at 70KB in memory to suit the computational constraints of a vision-based micro robot, the \textit{Colias}, while preserving key neural perception mechanisms. 
We further incorporated multi-attention mechanisms to emulate the distributed nature of LPLC2 responses, allowing the robot to detect and react to approaching targets both rapidly and selectively. 
We systematically evaluated the proposed method against a state-of-the-art locust-inspired collision detection model. 
Results showed that the fly-inspired visuomotor model achieved comparable robustness, at success rate of 96.1\% in collision detection while producing more adaptive and elegant evasive maneuvers. 
Beyond demonstrating an effective collision-avoidance strategy, this work highlights the potential of fly-inspired neural models for advancing research into collective behaviors in insect intelligence.

\end{abstract}

\section{Introduction}

Insects possess a remarkable ability to detect predators and escape swiftly, relying on complex yet highly efficient visual systems that operate with minimal neural resources \cite{1999_LGMD_Gabbiani}. 
This extraordinary capability has inspired efforts to endow robots with insect-inspired vision, enabling them to perceive and respond to threats in an agile and energy-efficient manner while using only modest computing power and inexpensive hardware. 
The central challenge is to design real-time, robust visual systems for artificial agents that remain effective under resource constraints \cite{2019_review_Fu}.

Locusts provide one of the earliest and most influential models for bio-inspired collision detection. 
Despite migrating in dense swarms \cite{bazazi2008collective}, they maintain individual safety through specialized visual neurons, such as the Lobula Giant Movement Detector (LGMD) \cite{o1974anatomy}. 
Over the past two decades, LGMD-inspired models have been developed into low-power, fast, and reliable visual strategies for robotic obstacle avoidance \cite{fu2019robust}. 
For example, Hu demonstrated the first implementation of an LGMD model on a compact ARM-based board, highlighting the feasibility of embedded vision for autonomous robots \cite{hu2016bio}. 
Later, Fu extended this line of work by integrating bilateral LGMD1 and LGMD2 neurons to enhance robustness in dynamic scenarios \cite{fu2017collision}. 
However, current LGMD-based approaches mainly provide global alerts to looming threats, which limits their ability to guide robots in executing direction-specific evasive maneuvers, thereby constraining safe and precise collision avoidance.

The fruit fly \emph{Drosophila} offers another powerful source of inspiration. 
Flies are notoriously difficult to catch because they can rapidly detect approaching threats and launch swift escape behaviors. More importantly, they evade selectively in the direction opposite the incoming object, ensuring both precision and effectiveness \cite{card2008visually}. 
This raises an important question: how can such fly-inspired visual control strategies be adapted to embedded robotic vision to enable not just collision detection, but also direction-selective and elegant evasive actions?

Recent research has identified the lobula plate/lobula columnar type-2 (LPLC2) neurons in \emph{Drosophila} as a critical component of this behavior. 
These neurons form a complex ensemble that responds strongly to outward motion from the receptive field center while being inhibited by inward motion, conferring an extreme selectivity for approaching threats \cite{2017_LPLC2_klapoetke,2022_LCs_klapoetke}. 
Motivated by this, numerous studies have sought to model and decode LPLC2 neural processing. 
Zhou et al. \cite{zhou2022shallow} employed machine learning techniques to capture its selective tuning properties. 
Hua et al. \cite{hua2022shaping} modeled LPLC2 with nonlinear computations and ON/OFF channel mechanisms. 
Zhao et al. \cite{zhao2023fly} advanced this work through population encoding and nonlinear integration, validating the model’s effectiveness in virtual robot experiments. 
Shuang et al. \cite{shuang2023opplod} further proposed an opponency-based looming detector that mimics LPLC2's synaptic processing with lateral inhibition and attention mechanisms. 
Most recently, Liu and Fu \cite{liu2025attention} introduced an attention-driven mLPLC2 model capable of detecting and localizing multiple approaching targets, offering promising insights for developing safer, direction-selective evasive strategies in robotics.

\begin{figure*}[h]
	\vspace{-10pt}
	\centering
	\includegraphics[width=\textwidth]{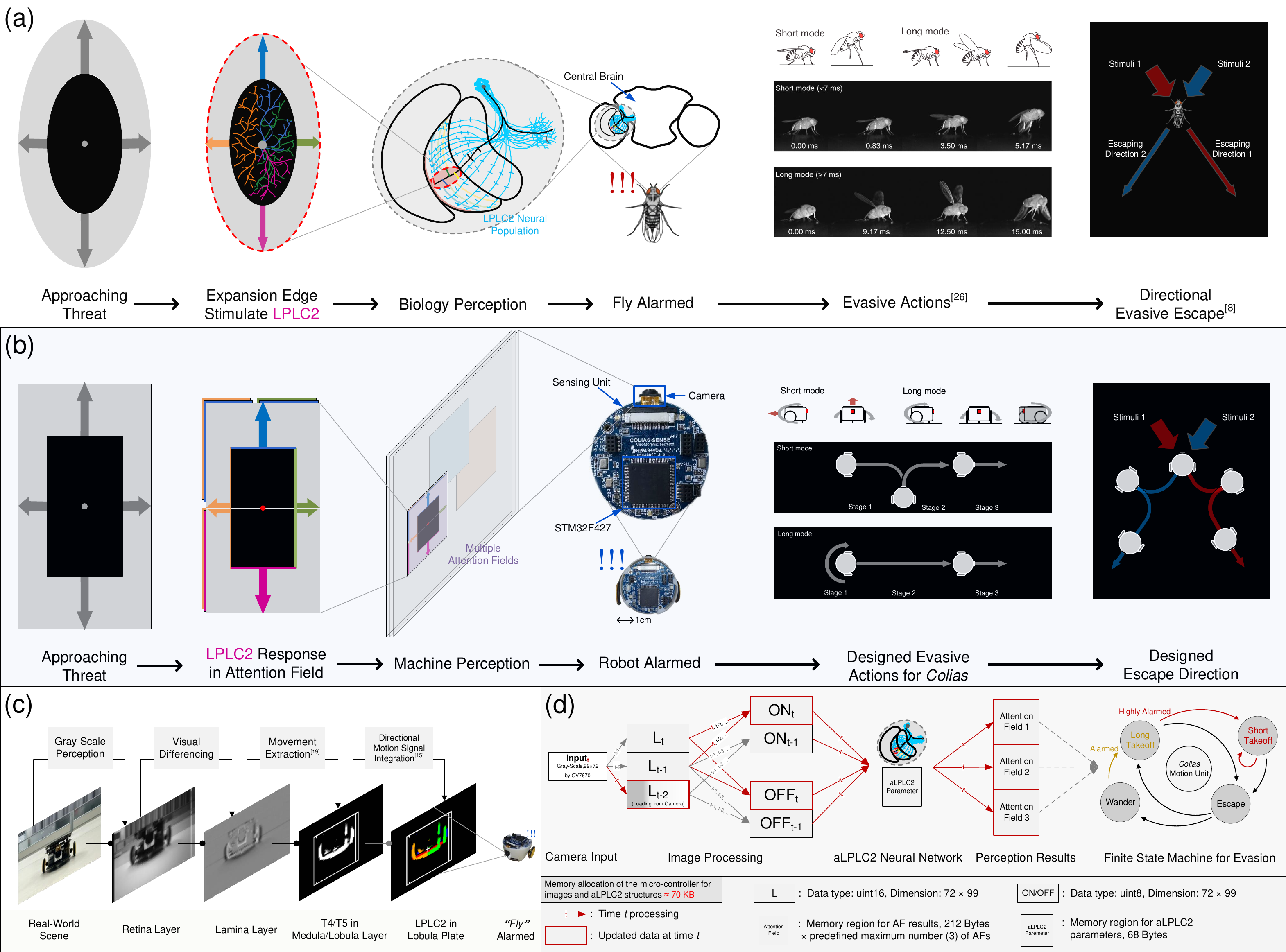}
	\caption{
		Illustrations of the visuomotor control of micro robots: (a) Fly biological LPLC2 visuomotor pathway.  
		\emph{Drosophila} can rapidly convert sensory cues into evasive actions to avoid predators. 
		The LPLC2 neurons, which are critical for this ability, not only distinguish approaching objects from other motion patterns but also form a lobula plate-spanning cluster \cite{2017_LPLC2_klapoetke}, providing global visual perception. 
		For evasive actions, fruit flies employ various escape strategies and can swiftly flee approximately 180$^\circ$ away from an approaching threat \cite{card2008visually}. (Image courtesy of \cite{2014_TAKEOFF}) 
		(b) Artificial attention-driven LPLC2 visuomotor system (ALVS). 
		We developed the attention-driven LPLC2 visuomotor system (ALVS), integrating visual perception and motion control. 
		After discretely designing and optimizing the aLPLC2 neural network, we deployed it on \textit{Colias}, a 4-cm-diameter ground micro-robot with highly constrained computational resources. 
		We also designed a motion system for the wheeled robot to enable rapid escape from threats.
		(c) Machine vision pipeline. 
		Visual signals are captured by the camera and processed layer by layer through the aLPLC2 neural network to trigger alertness in \textit{Colias}.
		(d) Signal storage and processing in the embedded system.
		A timing diagram illustrates signal updating and processing, along with a simple finite-state machine controlling escape behavior.
		The entire aLPLC2 feed-forward neural network is extremely compact, occupying only about 70 KB on the STM32F427 microprocessor.
		}
\label{fig2}
\vspace{-10pt}
\end{figure*}

In this paper, we investigate how fly collision-selective neural mechanisms can be leveraged to enhance robotic perception and motion in real-world settings. 
To this end, we developed the attention-driven LPLC2 Visuomotor System (ALVS), which integrates visual perception with motion control. 
The attention-driven LPLC2 (aLPLC2) neural network was designed and optimized in a discrete form and deployed on the \textit{Colias}, a ground micro-robot with a diameter of only 4-cm and severely limited computational resources \cite{Colias}. 
Equipped with this system, the robot was able to detect and localize multiple approaching threats and execute real-time, direction-selective evasive maneuvers toward the most salient danger—closely mirroring the collision-sensitive neural circuits of fly vision.

\section{The Proposed Robot Model}

In this section, we present the formulation of the Attention-LPLC2 neural network and its deployment on \textit{Colias}, a 4-cm-diameter micro-robot equipped with a low-power STM32F427 micro-chip.

\subsection{Visual Perception: Robot Model of Presynaptic Neuropils}
\label{Sec:Proposed Model}

The Attention-LPLC2 model is inspired by the biological architecture of the fruit fly \textit{Drosophila} visual system, which comprises multiple hierarchical processing layers \cite{2023_how_flies_see_motion}. 
The proposed visuomotor pathway can be broadly divided into three functional stages:
(1) the presynaptic neuropils, where luminance information is first processed to generate attention fields (AFs);
(2) the nonlinear integration of direction-selective signals in the AFs, producing ultra-selective responses to looming objects; and
(3) the convergence of visual signals onto premotor neurons, ultimately driving behavioral actions.
\\
The retina layer is derived from the compound eye structure of Drosophila, and is organized as a 72$\times$99 two-dimensional matrix of photoreceptors ($\mathit{P}$).
The retina layer ($R$) computes gray-scale brightness change at pixel location at $(x,y)$ as $R(x,y,t) = P(x,y,t) - P(x,y,t-\text{t}_i)$,
where $\text{t}_i$ is the time interval between two consecutive processing frames (33.33 ms) in the \textit{Colias} robot.
Subsequently, the signals was split into ON and OFF pathways, an essential phenomenon observed in biological visual systems \cite{2023_how_flies_see_motion,2010_ON/OFF_Drosophila,2020_fly_ON/OFF}. 
The ON pathway carries information about luminance increments, whereas the OFF pathway signals luminance decrements. This mechanism can be realized by `half-wave' rectifier which is defined as
\begin{equation}
	\label{Eq_ON/OFF}
	\begin{alignedat}{2}
		&L_{ON}(x,y,t)=w\cdot \frac{R(x,y,t) + \left|R(x,y,t)\right|}{2}, \\ 
		&L_{OFF}(x,y,t)=\frac{\big|\left|R(x,y,t)\right| - R(x,y,t)\big|}{2},
	\end{alignedat}
\end{equation}
where $w$ denotes the weight coefficient for the ON-channel signals.
Then, a 3$\times$3 mean-value filter is used to smooth the signal value in both the ON and OFF pathways for subsequent processing, which can be defined as: 
\begin{equation}
	\label{Eq_vDoG_convolution}
	\hat L_{ON(OFF)}(x,y,t)= \frac{1}{9} \cdot \sum_{i=-1}^{1} \sum_{j=-1}^{1} L_{ON(OFF)}(x+i,y+j,t),
\end{equation}

\begin{table}[!t]
	\caption{Parameter Configuration of aLPLC2 Neural Network}
	\centering
	\begin{tabular}{l l l}
		\hline
		Parameter & Description & Value\\ 
		\hline
		$w$ & the weight coefficient for ON-channel & 0.25 \\
		n & the number of connected neurons & 3 \\
		$R_{AF}$ & the radius of a single attention field & 36 \\
		$T_{a}$ & threshold for establishing attention field & 100 \\
		$T_{d}$ & threshold for conserving attention field & 5000 \\
		$\text{d}$ & time length for accumulating response  & 4 \\ 
		$w_s$ & the weight coefficient in speed calculation & 4000 \\
		$\alpha$ & the coefficient mapping position to angle & 0.707 \\
		$T_s$ & threshold for activating short-mode escape & 7000 \\
		\bottomrule
	\end{tabular}
	\vspace{-10pt}
	\label{Table_Params}
\end{table}

Next, we introduce the multiple-attention mechanism driven by bottom-up motion salience that equips the LPLC2 model with the capability for multi-target looming detection, forming the Attention-LPLC2 framework. 

At each time t, the bottom-up attention searches the visual field excluding regions covered by existing attention fields(AFs) for suspicious motion cues to generate a new AF.
The new attention field \(AF_{n+1} \doteq \{(x, y)\}\) is defined as a two-dimensional set of spatial coordinates in a square centered at the attention field centroid (AFC) \((x_{n+1}, y_{n+1})\), with a half-side length of \(L_{AF}\). 
Specifically, it selects \((x_{n+1}, y_{n+1})\) from all possible \((x, y)\) such that 
(i) it maximizes $\hat{L}_{\text{t}}(x,y) = \hat{L}_{\text{ON}}(x,y,\text{t}) + \hat{L}_{\text{OFF}}(x,y,\text{t})$, and 
(ii) the value \(\hat{L}_{\text{t}}(x,y)\) exceeds the detection threshold $T_a$.
Once selected, the new attention field \(AF_{n+1}(x_{n+1}, y_{n+1})\)
is incorporated into the set of all $N$ existing attention fields $\mathcal{AF}^{N}\doteq \{AF_0(x_0,y_0), \cdots , AF_n(x_n,y_n)\}$.
Accordingly, the whole process is defined as the following:
\begin{equation}
	\label{Eq_new_AF}
	\begin{aligned}
		&(x,y)\in \complement_\Omega \;\mathcal{AF}^{N}, \\
		(x_{n+1},&y_{n+1}) = arg\underset{x,y}{max}\big(\hat L_{\text{t}}(x,y)\big), \\ 
		\mathcal{AF}^{N\!+\!1}= \mathcal{AF}^{N}\!\cup &AF_{n+1}\!(x_{n\!+\!1},y_{n\!+\!1}), \text{if}\; \hat L_{\text{t}}(x_{n\!+\!1},y_{n\!+\!1}) \!>\! T_{a}. \\
	\end{aligned}
\end{equation}
$\Omega$ denotes for the entire 72$\times$99 visual field and $AF_0(x_0,y_0)\sim AF_{n+1}(x_{n+1},y_{n+1})$ represent all existing AFs.
As a next step, the centroid $(x_k, y_k)$ of each attention field $AF_k$ is updated by computing the intensity-weighted spatial moments of the \(L_{\text{t}}(x,y)\) matrix over the region defined by \(AF_k\). Specifically, the new centroid \((\bar{x}_k, \bar{y}_k)\) is obtained by calculating the normalized first-order moments:

\begin{equation} 
	\label{Eq_Update_Centroid_1}
	(\bar{x}_k, \bar{y}_k) =
	\left( 
	\frac{\sum_{AF_k}\!x\!\cdot\!L_{\text{t}}(x,y)}{\sum_{AF_k}\!L_{\text{t}}(x,y)},
	\frac{\sum_{AF_k}\!y\!\cdot\!L_{\text{t}}(x,y)}{\sum_{AF_k}\!L_{\text{t}}(x,y)}
	\right),
\end{equation}
where $k = 1, 2, \dots, N\!+\!1$ indicates the index of the existing AFs.
If the updated AFC $(\bar{x}_k, \bar{y}_k)$ lies within any other existing AFs, $AF_k$ will be discarded:
\begin{equation}
	\label{Eq_Fusion_Field}
	\mathcal{AF}^{N} \!=\! \mathcal{AF}^{N+1} \setminus AF_k,\;\text{if}\;(\bar{x}_k, \bar{y}_k)\in \complement_{\mathcal{AF}^{N\!+\!1}} AF_k,
\end{equation}
where $N$ denotes the number of existing attention fields.

In each existing AF, the directional selective information is nonlinearly integrated by a LPLC2 neuron centered at its AFC. 
The local motion information is computed using a Hassenstein-Reichardt Correlator (HRC) \cite{1956_Emd_HR} mimicking the direction-selective T4 cells (ON) in the medulla and T5 cells (OFF) in the lobula \cite{2013_T4/T5_ON/OFF,2023_how_flies_see_motion}.
The local motion at $(x,y)$ is denoted as $T_{4(5)}^\mathcal{V}(x, y, t)$, where $\mathcal{V}\in {\{r, l, d, u}\}$ represents rightward, leftward, downward and upward local motion, respectively, perceiving motion from one out of four cardinal directions in ON(OFF) channel \cite{1989_EMD_principles,2011_EMD_ON/OFF}. 
Taking the rightward motion detector $T_{4(5)}^r$ as an example here, each local pixel correlates and then sums input signals from $n$  multi-connected neurons:
\begin{equation}
	\label{Eq_EMD_HRC}
	\begin{aligned}
		T_{4}^r(x,y,t) = 
		\sum_{c=1}^{\text{n}} \big( L_{\text{ON}}(x&+c\cdot\text{s},y,t) \cdot L_{\text{ON}}(x,y,t-C_d\cdot\text{t}_i) \\
		\quad -L_{\text{ON}}&(x,y,t)\cdot L_{\text{ON}}(x+c\cdot\text{s},y,t) \big)
	\end{aligned}
\end{equation}
where $C_d$ is a coefficient for time delay, n indicating the number of the neuron connected to $T_{4(5)}^r$ at $(x,y)$, and \text{s} is the sampling distance increment. Then, the calculation of local motion $(LM^\mathcal{V})$ is completed by the convergence of the signals from ON and OFF channels, that is:
\begin{equation} 
	\label{Eq_Local_Motion_4_direction}
	\begin{aligned}
		LM_r(x,y, t)=T_{4}^r(x,y,t)+T_{5}^r(x,y,t).
	\end{aligned}
\end{equation}

Next, directional signals covered by each existing AF is integrated in a highly nonlinear manner. 
Specifically for a single LPLC2 cell at the center of $AF_k$, each arm of its cross-shaped primary dendrites ramifies in one of the lobula plate layers and extends along that layer's preferred motion direction \cite{2017_LPLC2_klapoetke}. 
Within $AF_k$, the motions toward four cardinal directions are mapped into four quadrants ($Q^{1\sim4}_k$):
\begin{equation}
	\label{Eq_Integration_4_Quadrants}
	\begin{aligned} 
		&Q_k^1(t) = \!\textstyle \sum_{\Gamma_{k}^1}\,\! \big(LM_r(u,v,t) + LM_u(u,v,t)\big); \\
		&Q_k^2(t) = \!\textstyle \sum_{\Gamma_{k}^2}\,\! \big(LM_l(u,v,t) + LM_u(u,v,t)\big); \\
		&Q_k^3(t) = \!\textstyle \sum_{\Gamma_{k}^3}\,\! \big(LM_l(u,v,t) + LM_d(u,v,t)\big); \\
		&Q_k^4(t) = \!\textstyle \sum_{\Gamma_{k}^4}\,\! \big(LM_r(u,v,t) + LM_d(u,v,t)\big),
	\end{aligned}
\end{equation}
where $\Gamma_{k}^1 \!\sim\! \Gamma_{k}^4$ denote the sets of coordinates in the four quadrants of the $AF_k$, divided by its AFC located at $(\bar{x_k},\bar{y_k})$.

\begin{figure*}[!t]
	\vspace{-10pt}
	\centering
	\includegraphics[width=\textwidth]{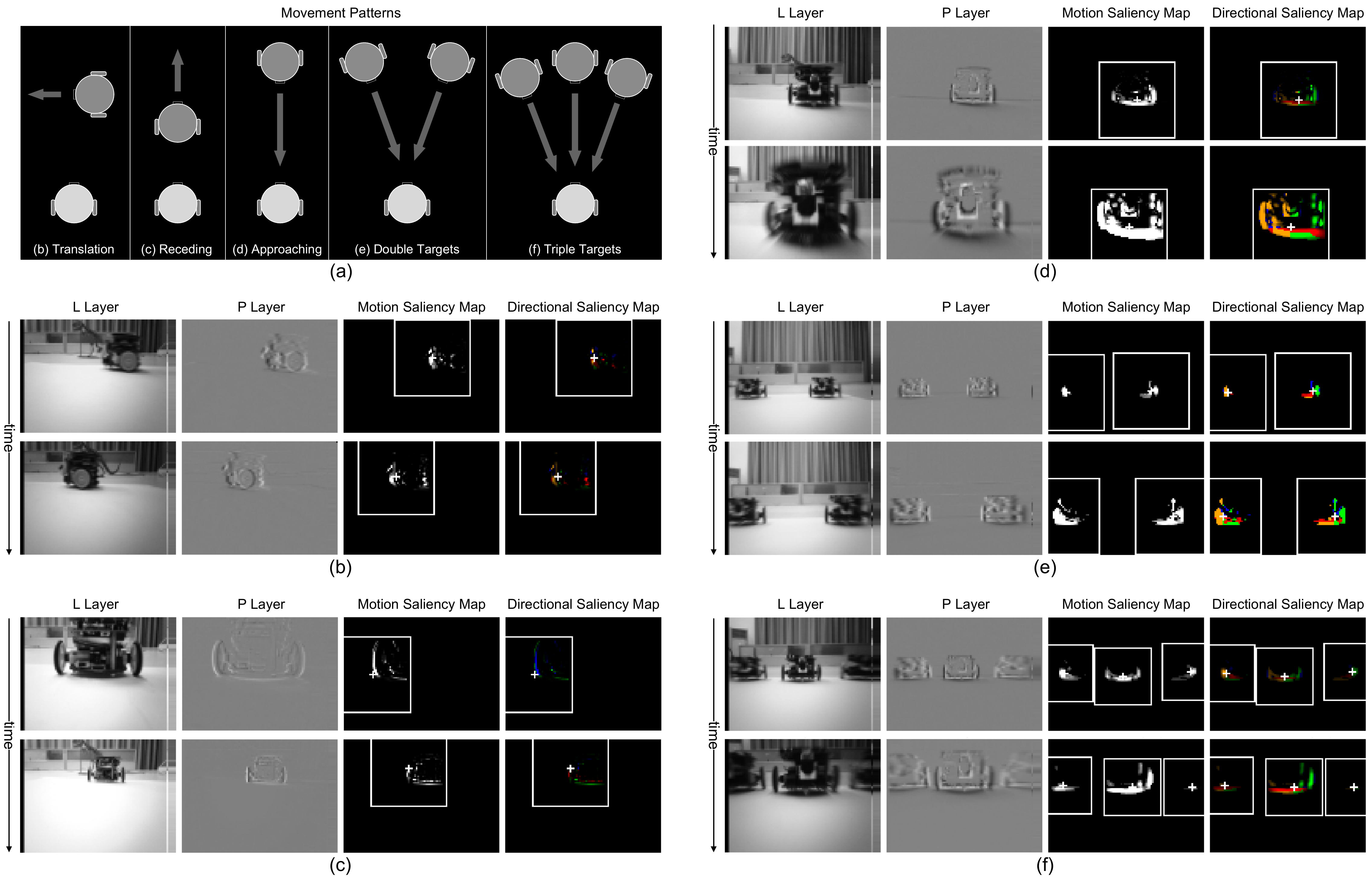}
	\caption{
		Visualization of the intermediate processing in the visual-perception submodule, where (a) illustrates five representative basic test scenarios and (b) shows the corresponding outputs from four intermediate layers: L layer, P layer, motion saliency map, and directional saliency maps.
		}
	\label{fig3}
	\vspace{-10pt}
\end{figure*}

Only when all four dendritic arbors have received their preferred motion signal from $AF_k$, its postsynaptic $LPLC2_k(t)$ will be activated. 
The attention-driven regional information is integrated as
\begin{equation}
	\label{Eq_Integration_LPLC2_3}
	\begin{aligned}
		LPLC2_k(t)&=bool\Big( {\textstyle \prod_{i=1}^{4} Q_k^i(t) } \Big) \\
		&\times \big(Q_k^1(t)+Q_k^2(t)+Q_k^3(t)+Q_k^4(t)\big),
	\end{aligned}
\end{equation}
where $bool(\cdot)$ function outputs $1$ once none of $Q_1 \!\sim\! Q_4$ equals $0$. 

Importantly, if the LPLC2 cell in current $AF_k$ remains inactive or shows minimal activation, the $AF_k$ will disappear, mathematically expressed as follows:
\begin{equation}
	\label{Eq_deField}
	\mathcal{AF}^{N-1}\!=\! \mathcal{AF}^{N} \setminus AF_k,\;\text{if}\;\frac{1}{\text{d}}\cdot\!\sum^t_{m=t-\text{d}}\!LPLC2_k(m) < T_{d},
\end{equation}
where $T_{d}$ is a threshold and d denotes the time duration coefficient over which responses are accumulated within the current AF. 
However, the Attention-LPLC2 model ensures that at least one AF is always retained.

\begin{figure*}[!t]
	\vspace{-10pt}
	\centering
	\includegraphics[width=\textwidth]{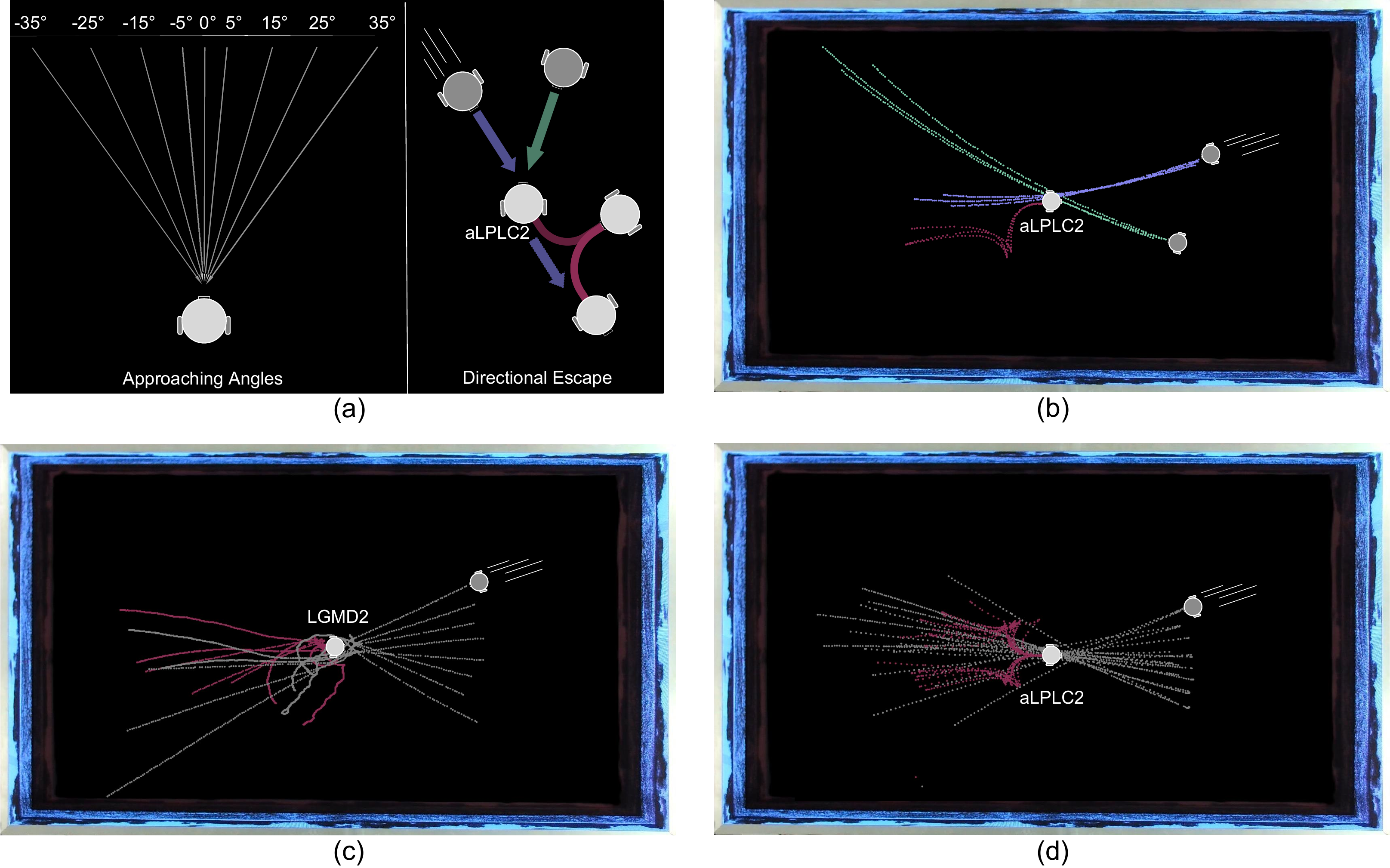}
	\caption{
		Trajectory demonstrations of directional escape behaviors. (a) Experimental setup with a single robot approaching from different angles, and motion control of the tested robot under multiple approaching threats. (b) Visualization of the trajectories of the tested robot and two simultaneously approaching robots. (c) and (d) illustrate the escape trajectories of the LGMD2-based system and the aLPLC2-based system, respectively.
	}
	\label{fig4}
	\vspace{-10pt}
\end{figure*}

\subsection{The Visuomotor Pathway}

The separate neural signals from vision system ultimately come together to coherently influence action \cite{2023_Visuomotor,2017_Feature}.
Once there is any AF being stimulated, the visuomotor pathway converge colliding information from $\mathcal{AF}^N$ to drive action.
The premotor neurons sum response from all existing AFs as the input strength, and then convert it to the escaping speed as
\begin{equation}
	\label{Eq_Strength}
	S(t) = 1/\big(1 + e^{-\frac{1}{w_s}\sum_{k=1}^{N}LPLC2_k(t)}\big),
\end{equation}
where $w_s$ is a weight coefficient, and $N$ denotes the number of currently existing AFs. 

Benefit from the attention mechanism, \textit{Colias} has the ability to spatially localize a salient stimulus and act accordingly, as observed in the fruit fly \textit{Drosophila} \cite{2008_Escape_Direction}. 
The escaping direction is determined by the horizontal centroid $\bar{X}$ of all LPLC2 response strengths across existing $N$ AFs within the 99-pixel horizon.
The horizontal centroid $\bar{X}$ of global threat at time $t$ is estimated by
\begin{equation}
	\bar{X}(t)
	\;=\;
	\frac{\sum_{k=1}^{N}\bar{x_k}\!\cdot \!LPLC2_k(t)}
	{\sum_{k=1}^{N}LPLC2_k(t)}
\end{equation}
supposing each $AF_k$ contributes responses $LPLC2_k(t)$ at $(\bar{x_k},\bar{y_k})$.
When confronted by a predator-mimicking looming stimulus, a fly responds with either a long-duration escape behavior sequence that initiates stable flight or a distinct, short-duration sequence that sacrifices flight stability for speed \cite{2014_TAKEOFF}. 
Before takeoff, flies begin aseries of postural adjustments that determine the direction of their escape and then jump away from the expanding visual stimulus\cite{2008_Escape_Direction}.
We assume \textit{Colias}'s forward direction is $0^\circ$ and rear is $-180^\circ$; left turns are negative angles, right turns positive.
In the \textit{long takeoff} state, \textit{Colias} will firstly attempt to turn $180^\circ$ away from $\bar X(t)$ as:
\begin{equation}
	\label{Eq_Direction_Long}
	\theta_{long}(t)=\Big(\alpha\! \cdot \!\big(\bar X(t)-49\big)\Big)^\circ+180^\circ,
\end{equation}
where $\alpha$ is the coefficient mapping horizontal position in $[-49,49]$ to angle in $[-35^\circ,35^\circ]$, corresponding to the horizontal visual field of \textit{Colias} \cite{Colias}. 
After gesture adjustment by this turning, \textit{Colais} moves into the \textit{escape} state and escapes straight forward. 
However, the gesture adjustment could be interrupted by a fierce stimuli \cite{2014_TAKEOFF}.

When the strength of the global threat exceeds the threshold $Ts$, the current state is overridden by the \textit{short takeoff} state. 
To rapidly and effectively evade an approaching object, \textit{Colias} initiates a two-stage escape response upon activation of short-mode:  
(1) a swift retreat combined with turning, terminating when its heading deviates by $90^\circ$ from the threat direction that triggered the short-mode;  
(2) forward movement away from the threat while continuing to turn, ending when it faces $180^\circ$ opposite the threat.  
The escape direction among these two phases $\theta^1_{short},\theta^2_{short}$ is calculated as: 
\begin{equation}
	\label{Eq_Direction_Short_1}
	\theta_{short\!}^1(t)\!=\!
	\left\{
	\begin{aligned}
		&\!\Big(\!\alpha\! \cdot \!\big(\!\bar X(t)-49\!\big)\!\Big)^\circ\!-\!90^\circ\!, \text{if}\;\bar X(t)\!-\!49\!\ge\! 0\\
		&\!\Big(\!\alpha\! \cdot \!\big(\!\bar X(t)-49\!\big)\!\Big)^\circ\!+\!90^\circ\!, \text{if}\;\bar X(t)\!-\!49\!<\! 0\,,
	\end{aligned}
	\right.
\end{equation}
\begin{equation}
	\label{Eq_Direction_Short_2}
	\theta_{short}^2(t)=
	\left\{
	\begin{aligned}
		&\theta_{short}^1(t)-90^\circ, \text{if}\;\theta_{short}^1(t)\!<\! 0\\
		&\theta_{short}^1(t)+90^\circ, \text{if}\;\theta_{short}^1(t)\!\ge\! 0\,,
	\end{aligned}
	\right.
\end{equation}
The overall trajectory of these two phases forms a $\gamma$-shaped path, ultimately orienting \textit{Colias} $180^\circ$ opposite to the threat. Once escaped successfully, \textit{Colias} switches to the \textit{wander} state, moving forward in a constant velocity.


\begin{figure*}[!t]
	\vspace{-10pt}
	\centering
	\includegraphics[width=0.9\textwidth]{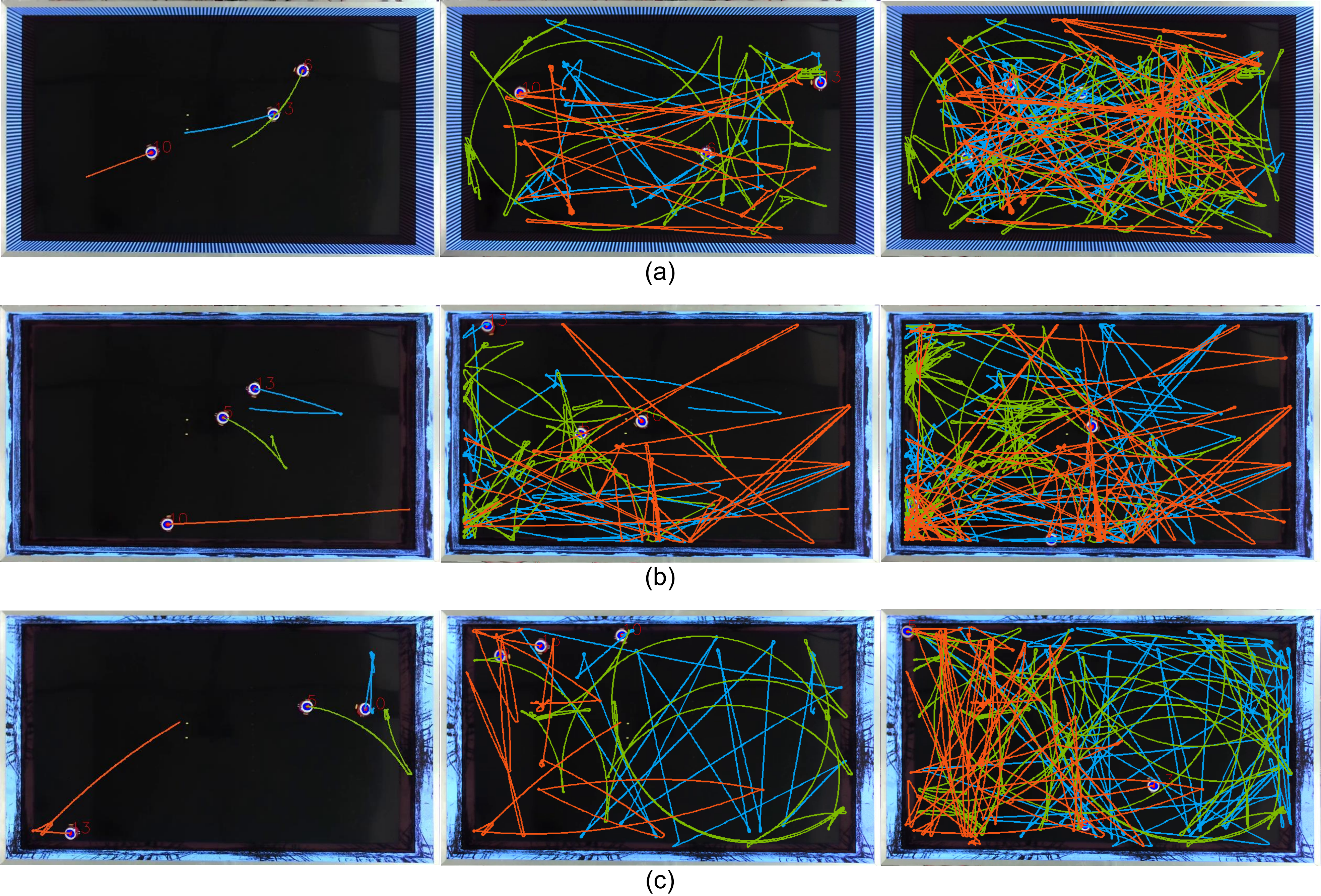}
	\caption{
		Navigation dynamics of the comparative LGMD robot model with three robots operating simultaneously under three different visual backgrounds.
	}
	\label{fig5}
	\vspace{-10pt}
\end{figure*}

\subsection{Embodiment: Visuomotor Control of Micro Robot}

The experiments in this study were conducted on the \textit{Colias} micro-robot platform \cite{Colias}, where the proposed fly-inspired visual system was deployed for real-time visual processing under constrained computational resources. 
At the core of its Central Sensor Unit (CSU) is an ARM Cortex-M4F micro-controller, equipped with a hardware floating-point unit to support efficient on-board execution of the neural model.

In this research, robot relied solely on vision as its sensing modality. 
Visual input was provided by a low-power OV7670 CMOS image sensor with a horizontal field of view of approximately $70^{\circ}$, capable of capturing up to 30 frames per second (fps) at VGA resolution (640$\times$480). 
To balance image quality with processing efficiency, the system operated at a reduced resolution of 72$\times$99 pixels while maintaining a real-time frame rate of 30 fps. The camera output was formatted in 8-bit YUV422, separating luminance (Y) and chrominance (UV) channels. 
Since the Attention-LPLC2 model processes only grayscale input, the Y channel was used directly. 
This design significantly reduced memory usage and computational overhead, enabling stable real-time processing on the resource-limited embedded platform.

For closed-loop experiments, the motion states of robots were governed by a finite state machine (FSM) \cite{1996_FSM} with four states, completing a full wander–escape behavioral cycle.

\section{Experimental Evaluation}

In this section, we present a series of systematic experiments on the \textit{Colias} micro-robot, covering the full process from collision detection to directional escape. 
First, we evaluated the visual-perception submodule to assess its ability to localize multiple approaching targets. 
Second, we conducted both open-loop and closed-loop comparative tests to examine the robustness and effectiveness of the proposed embedded visuomotor system in collision avoidance. 
For benchmarking, a state-of-the-art LGMD2-based embedded system \cite{fu2019robust} was adopted as the baseline model.

\begin{figure*}[!t]
	\vspace{-10pt}
	\centering
	\includegraphics[width=0.9\textwidth]{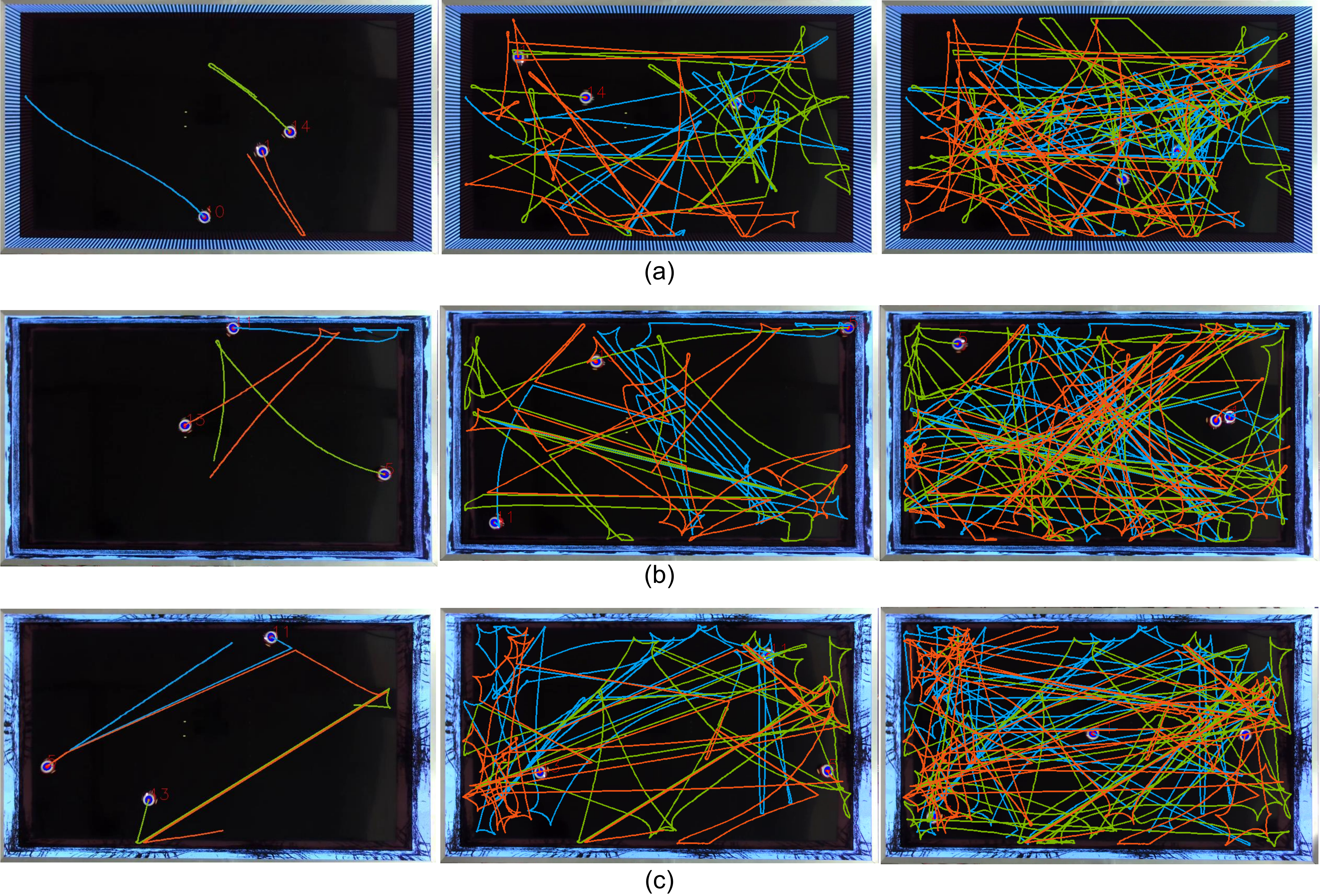}
	\caption{
		Navigation dynamics of three ALVS agents in an arena across three distinct visual backgrounds.
	}
	\label{fig6}
\end{figure*}

\begin{figure*}[!t]
	\centering
	\includegraphics[width=0.7\textwidth]{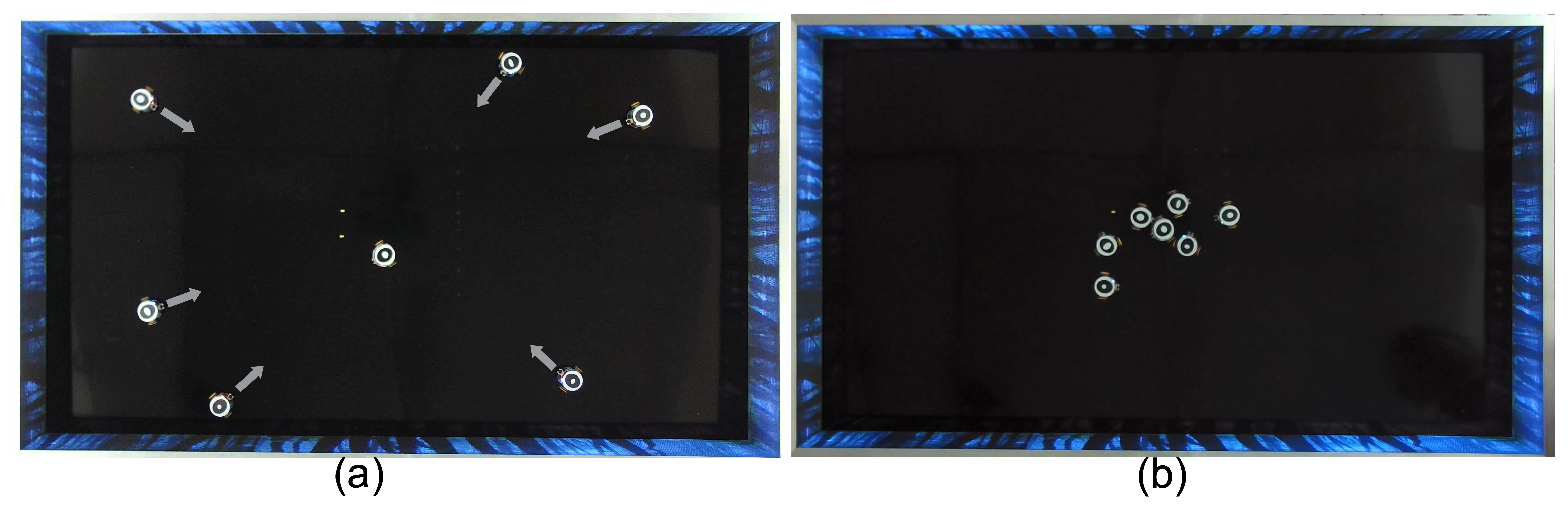}
	\caption{
		A simple swarming scenario achieved with the modified ALVS, in which six micro-robots moved toward a swarm leader and eventually clustered together.
	}
	\label{fig7}
	\vspace{-10pt}
\end{figure*}

\subsection{Multi-Attention and Approaching-Detection}

In the first set of experiments, the robot equipped with ALVS was kept stationary, while an obstacle robot moved in five distinct patterns to provide diverse visual stimuli: translation, receding, approaching, double approaching, and triple approaching targets. During each trial, we recorded the output of the visual-perception submodule and selected two representative frames to illustrate its four key processing layers.

The visualizations of these five scenarios are presented in Figure~\ref{fig3}. When encountering multiple moving targets, the visual-perception subsystem successfully generated a corresponding number of attention fields and precisely localized the center coordinates of each field (Figure~\ref{fig3}~(e)–(f)), thereby enabling localized attention to every moving object within the visual field. In addition, the directional saliency maps—characterized by four expansion edges resembling the dendritic arbors of LPLC2 neurons (Figure~\ref{fig2})—showed clear distinctions between motion types: more dilation edges were detected in approaching scenarios, whereas fewer appeared in non-approaching cases (Figure~\ref{fig3}~(b)–(c)).

Overall, these results confirm that the visual-perception sub-module provides robust multi-target localization and collision detection, delivering accurate feed-forward signals for the subsequent directional escape control within ALVS.

\subsection{Directional Evasion}

This open-loop experiment examined the advantages of incorporating multiple attention mechanisms in enabling the robot to perform direction-selective escape behaviors against threats approaching from different directions.

In the first set of trials, the tested robot was placed at the same initial position, while an obstacle robot advanced toward it at a constant velocity from nine angles ranging from $-35^\circ$ to $35^\circ$ (Figure~\ref{fig4}~(a)). The trajectory results showed that, regardless of the approach angle, our embedded system with multi-attention mechanisms consistently selected the correct escape direction (Figure~\ref{fig4}~(d)). By contrast, the comparative system, lacking directional information, failed to execute direction-selective evasions, resulting in a higher likelihood of collisions during motion control (Figure~\ref{fig4}~(c)).

In the second part, the experimental challenge was increased by introducing two obstacle robots that approached simultaneously from different directions at constant but unequal speeds. As shown in Figure~\ref{fig4}~(b), ALVS identified the faster of the two incoming robots (blue trajectory) as the more imminent threat and executed an evasion maneuver opposite to its approach. This strategy enabled the tested robot to move safely away from the most conspicuous danger, thereby enhancing collision avoidance under multi-threat scenarios.

\subsection{Arena Navigation Dynamics}

To validate the robustness and effectiveness of the proposed closed-loop visuomotor system, we conducted comparative tests against an LGMD2-based embedded system in an arena environment. In each trial, three robots were deployed to create a multi-target scenario. All robots were equipped with the same embedded system (either ALVS or LGMD2-based), moved simultaneously at a constant velocity of approximately 10 cm/s, and executed the corresponding collision avoidance strategy. Each trial lasted 10 minutes. To assess generalizability, both models were tested under three diverse visual scenes, including two complex outdoor environments. Robot trajectories were recorded by a top-down camera, with individual robots distinguished by unique patterns \cite{krajnik2013external}. Representative results are shown in Figure~\ref{fig5}. 
The comparative arena dynamics of LGMD2-based agents is shown in Figure~\ref{fig6}.

Based on the data analysis, a collision event was defined when the distance between two robots, or between a robot and the arena wall, fell below the robot’s diameter. The success rates of the LGMD-2 and LPLC2 models were 93.7\% and 96.1\%, respectively. For robots guided by the LGMD-2 model, most collisions occurred during interactions between robots, whereas for LPLC2-guided agents, collisions were more often with the arena boundaries.

\subsection{Towards Insect Collective Behavior}

From the above experiments, we conclude that the proposed model not only detects multiple approaching objects effectively but also demonstrates robust collision avoidance capabilities. This raises an important question: could such an algorithm also contribute to swarm intelligence systems?

By leveraging its multi-object detection ability, the robot can rapidly react and take appropriate actions when encountering multiple approaching agents. To explore this capability, we programmed the robot to perform an automatic spinning action upon detecting an approaching object, rather than executing an escape maneuver. This behavior simulates a basic swarm interaction, where individuals in nature perceive one another and respond accordingly (Figure~\ref{fig7}).\footnote{The full demonstration is provided in the supplementary video.}

The results suggest that this algorithm, by enabling broader environmental perception, can enhance the responsiveness of swarm intelligence systems to dynamic environments and thereby promote more effective collective behaviors within robot swarms.

\section{Concluding Remarks}

In this work, we introduced a fly-inspired visuomotor control strategy for collision perception and reactive evasion in micro-robots. 
By modeling the LPLC2 neural circuits and integrating multi-attention mechanisms, the proposed system enabled the \textit{Colias} micro-mobile robot to detect and localize multiple approaching threats and to perform real-time, direction-selective evasive maneuvers. 
Experimental results showed that our method achieved collision detection robustness comparable to a state-of-the-art locust-inspired model, while producing more adaptive and efficient escape behaviors. 
Beyond validating the effectiveness of LPLC2-inspired neural mechanisms on a resource-constrained robotic platform, this study highlights their potential for advancing collective behaviors research via mimicking insect intelligence. 
Future work will extend this framework to multi-robot coordination, incorporate richer sensory modalities, and explore adaptive learning strategies to further enhance the scalability and versatility of the proposed fly-inspired visuomotor control.


\end{document}